\documentclass[10pt,twocolumn,letterpaper]{article}

\usepackage{cvpr}              % To produce the CAMERA-READY version
\usepackage{multirow}

\usepackage{xcolor}  % 提供颜色支持
\usepackage{colortbl}  % 提供表格颜色支持
\usepackage{booktabs} %表格中toprule等需要，aaai默认没
\usepackage{bm}

\usepackage{subcaption}
\usepackage{amsmath}
\usepackage{amssymb}
\definecolor{cvprblue}{rgb}{0.21,0.49,0.74}
\usepackage[pagebackref,breaklinks,colorlinks,allcolors=cvprblue]{hyperref}

\def\paperID{186} % *** Enter the Paper ID here
\def\confName{CVPR}
\def\confYear{2026}

\title{An Instance-Aware Prompting Framework for Training-free Camouflaged Object Segmentation}

\author{
Chao Yin,
Jide Li,
Hang Yao,
Xiaoqiang Li \\
Shanghai University, Shanghai, China\\
{\tt\small yincaho@shu.edu.cn, iavtvai@shu.edu.cn, yaohang@shu.edu.cn, xqli@shu.edu.cn}
}

\begin{document}
\maketitle
\begin{abstract}
% Training-free Camouflaged Object Segmentation (COS) seeks to segment camouflaged objects without task-specific training, by automatically generating visual prompts to guide the Segment Anything Model (SAM). However, existing pipelines mostly yield semantic-level prompts, which drive SAM to coarse semantic masks and often collapse multiple camouflaged instances into one region. To address this critical limitation, we propose an \textbf{I}nstance-\textbf{A}ware \textbf{P}rompting \textbf{F}ramework (IAPF) for training-free COS that upgrades prompt granularity from semantic to instance-level while keeping all components frozen. Specifically, the IAPF comprises three steps: (1) Text Prompt Generator, prompting a Multimodal Large Language Model (MLLM) for producing image-specific foreground/background tags; (2) \textbf{Instance Mask Generator}, leveraging a detector-agnostic box enumerator to produce precise instance-level box prompts, alongside the proposed Single-Foreground Multi-Background Prompting strategy to sample region-constrained point prompts within each box, enabling SAM to yield a candidate instance mask; (3) Self-consistency Instance Mask Voting, which selects the final COS prediction by identifying the candidate mask most consistent across multiple candidate instance masks. Extensive evaluations on three challenging COS benchmarks, two Camouflaged Instance Segmentation(CIS) benchmarks, and two downstream application datasets demonstrate that IAPF achieves state-of-the-art performance among training-free methods. Code will be released upon acceptance.

Training-free Camouflaged Object Segmentation (COS) seeks to segment camouflaged objects without task-specific training, by automatically generating visual prompts to guide the Segment Anything Model (SAM). However, existing pipelines mostly yield semantic-level prompts, which drive SAM to coarse semantic masks and struggle to handle multiple discrete camouflaged instances effectively. To address this critical limitation, we propose an \textbf{I}nstance-\textbf{A}ware \textbf{P}rompting \textbf{F}ramework (IAPF) tailored for the first training-free COS that upgrades prompt granularity from semantic to instance-level while keeping all components frozen. The centerpiece is an Instance Mask Generator that (i) leverages a detector-agnostic enumerator to produce precise instance-level box prompts for the foreground tag, and (ii) introduces the Single-Foreground Multi-Background Prompting (SFMBP) strategy to sample region-constrained point prompts within each box prompt, enabling SAM to output instance masks. The pipeline is supported by a simple text prompt generator that produces image-specific tags and a self-consistency vote across synonymous task-generic prompts to stabilize inference. Extensive evaluations on three COS benchmarks, two CIS benchmarks, and two downstream datasets demonstrate state-of-the-art performance among training-free methods. Code will be released upon acceptance.
\end{abstract}    
\section{Introduction}
\label{sec:intro}

\begin{figure*}[t]
  \centering
  \begin{subfigure}[b]{0.60\linewidth}
    \includegraphics[width=1.0\linewidth]{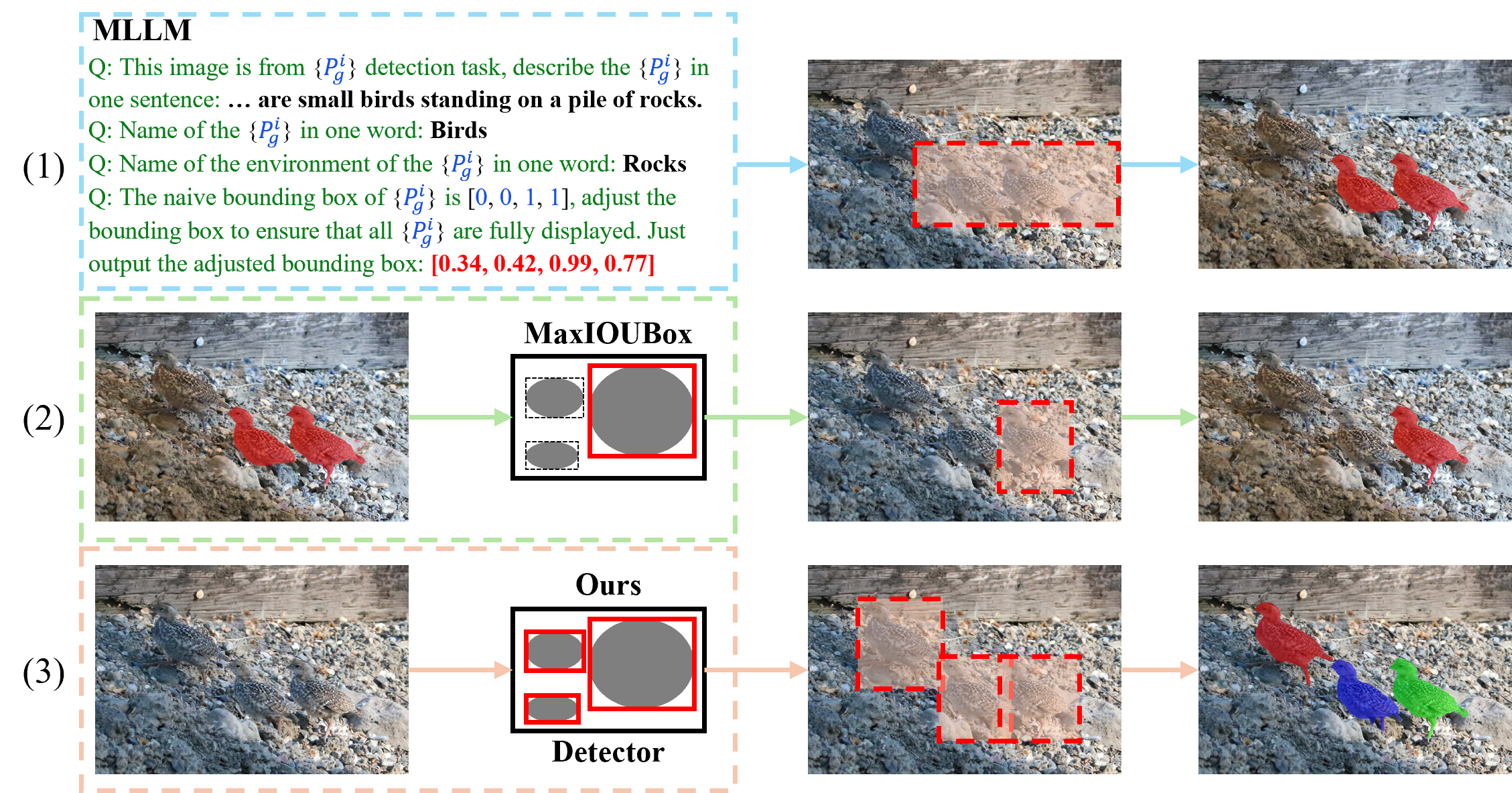}
    \caption{}
    \label{fig: motivationFigure}
  \end{subfigure}
  \hfill
  \begin{subfigure}[b]{0.35\linewidth}
    \includegraphics[width=1.0\linewidth]{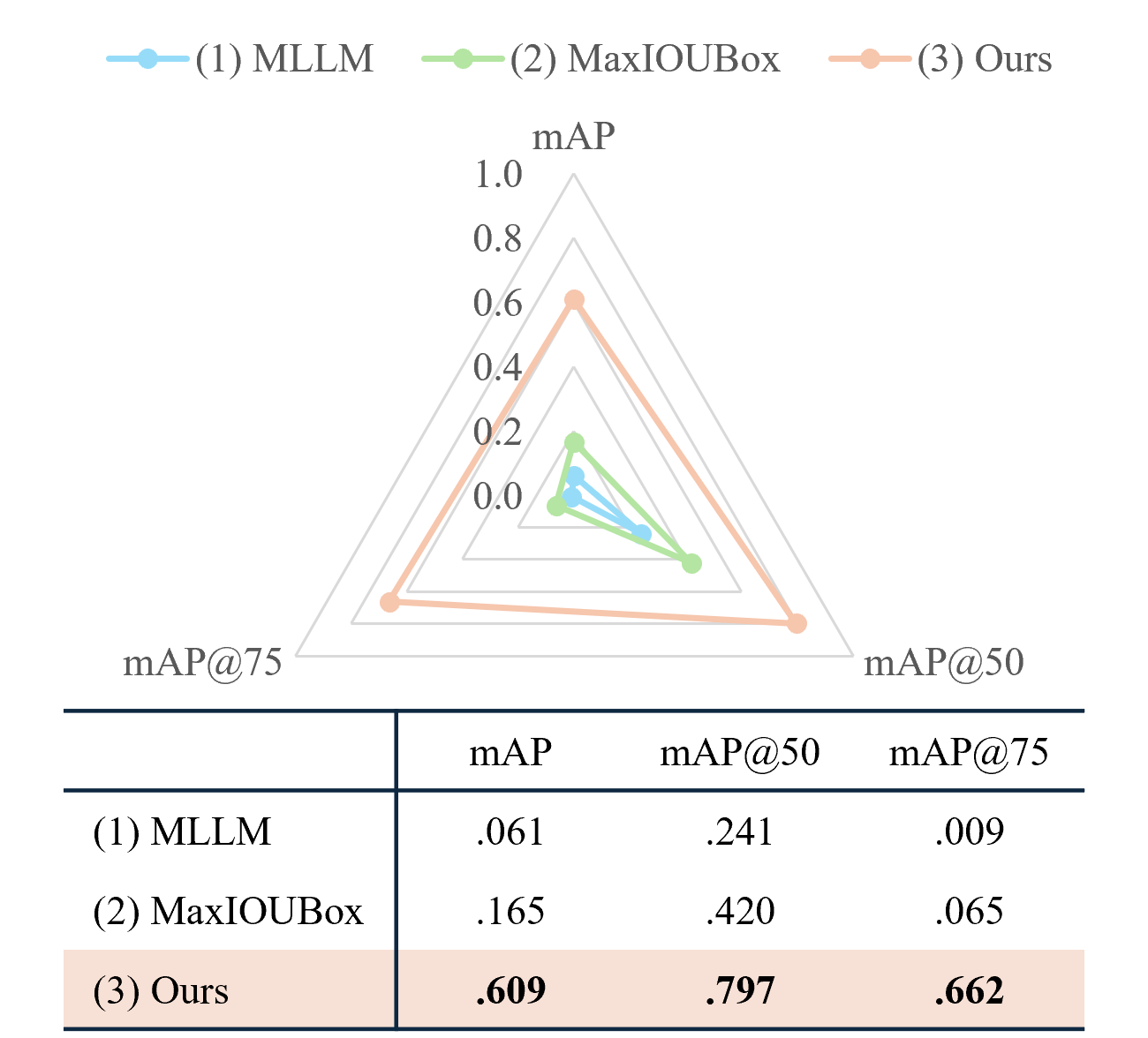}
    \caption{}
    \label{fig: motivationTable}
  \end{subfigure}
  \caption{Motivation of the proposed IAPF. (a) Visualization of box prompts generated by MLLM, MaxIOUBox, and our method. Only the instance-aware strategy yields multiple instance-level boxes. (b) Quantitative results of the COD10K dataset showing our superior box prompts accuracy, especially under a high IoU threshold.}
  \label{fig: motivation}
\end{figure*}

Camouflaged Object Segmentation (COS) holds significant research value and broad application prospects, including agriculture~\cite{Yang2025,wang2024depth}, medicine~\cite{fan2020inf}, and military reconnaissance~\cite{truong2024style}, yet remains challenging because targets are visually indistinguishable from their surroundings. Fully-supervised COS methods~\cite{fan2020camouflaged,yin2024camouflaged} heavily rely on dense pixel annotations, which are costly and limit practical deployment. Weakly-supervised and unsupervised alternatives~\cite{chen2024sam,du2025shift} mitigate labeling, but still degrade as supervision becomes sparse.

Recent advancements in promptable segmentation models (\eg, Segment Anything Model (SAM)~\cite{kirillov2023segment}) have demonstrated promising potential, achieving competitive results across various segmentation tasks with minimal manual visual prompts (\ie, points and boxes). Nevertheless, manual prompts significantly constrain their practical deployment when segmenting specific objects of interest. Consequently, increasing attention has been devoted to developing training-free segmentation methods that automatically generate visual prompts without additional supervision. An active research area within this context is training-free COS methods~\cite{yin2025stepwise,hu2024relax,hu2024leveraging,tang2024chain}, which utilize a single task-generic prompt (\eg, "\textit{camouflaged animal}") indiscriminately applied across all test samples within the COS domain to automatically generate visual prompts required by SAM, thus enabling segmentation of camouflaged objects of interest. 

Despite such advancements, existing training-free COS pipelines still generate semantic-level visual prompts rather than instance-level ones, which constrains SAM to output coarse semantic masks and prevents reliable handling of multiple discrete camouflaged objects. According to SAM, explicit bounding boxes are required for instance segmentation; therefore, enumerating instance-level boxes is essential. However, current practices either ask an MLLM to return a single coarse box via VQA (\Cref{fig: motivationFigure}(1)) or derive a single box from an intermediate semantic mask through MaxIOUBox (\Cref{fig: motivationFigure}(2)), both collapsing multiple instances into one region and failing under multi-object camouflage.

Considering the above issues, we adopt an instance-aware prompting strategy: enumerate all camouflaged instances with multiple boxes and constrain point prompting per box to steer SAM from semantic to instance outputs. The box enumerator is detector-agnostic (instantiated with Grounding DINO~\cite{liu2024grounding}) and serves only to provide instance-level box prompts (\Cref{fig: motivationFigure}(3)). As shown in \Cref{fig: motivationTable}, our box prompts achieve mAP $0.662$@ IoU $0.75$ on COD10K, far surpassing semantic-level baselines ($0.009$, $0.065$).

\begin{figure*}[t]
\centering
\includegraphics[width=0.9\textwidth]{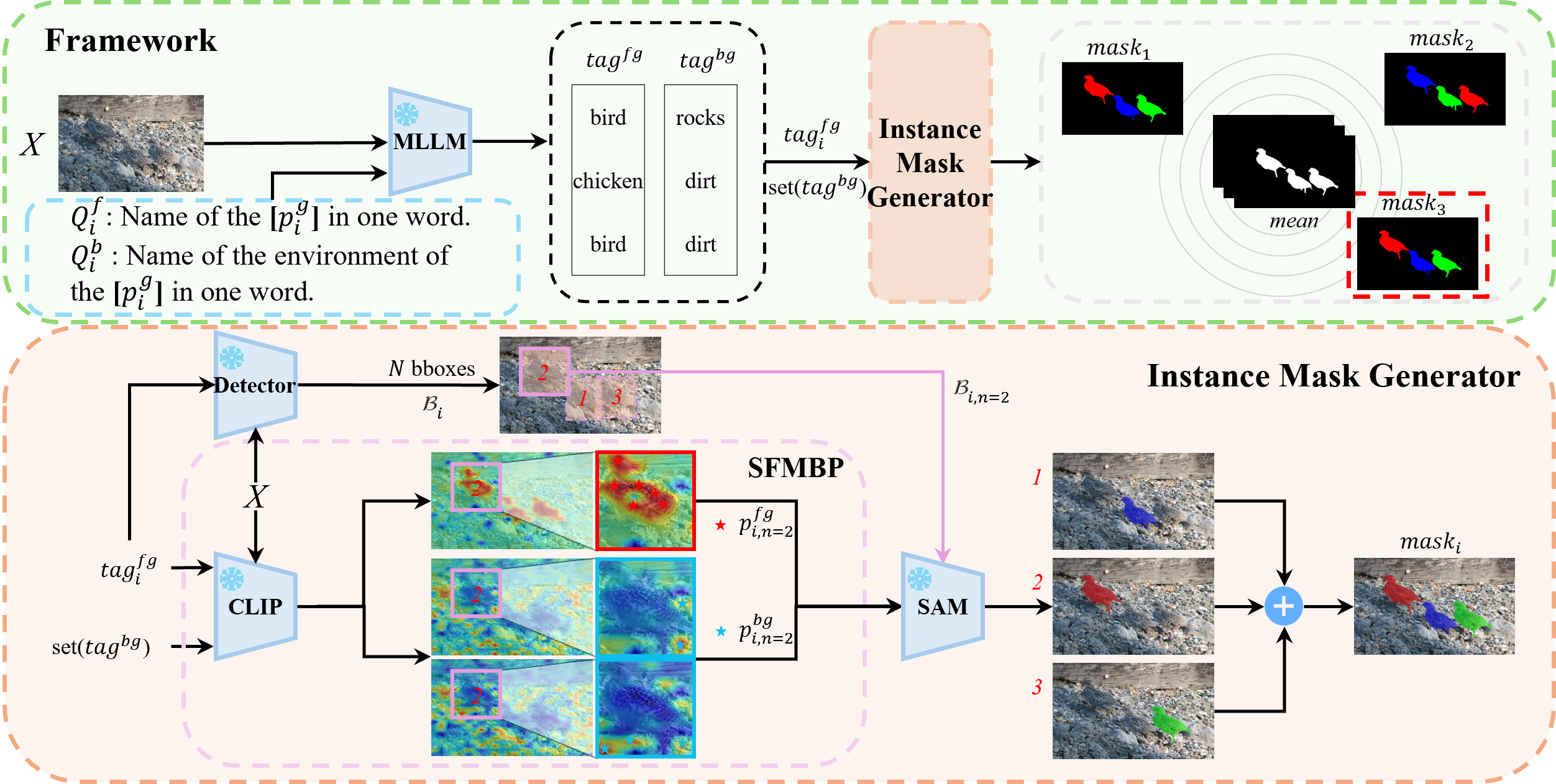} % Reduce the figure size so that it is slightly narrower than the column.
\caption{Framework of the proposed IAPF, which consists of three steps:
(1) \textbf{Text Prompt Generator}: an MLLM turns task-generic prompts and an image into foreground/background tags.
(2) \textbf{Instance Mask Generator} (bottom): a detector-agnostic enumerator produces $N$ instance-level box prompts; SFMBP uses CLIP to form foreground/multi-background heatmaps and samples region-constrained per-box points; SAM converts $N$ box–points pairs into a candidate instance mask.
(3) \textbf{Self-consistency Instance Mask Voting}: the candidate whose semantic projection is most consistent across repetitions is selected as the final COS prediction (its instance masks provide CIS).}
\label{fig: framework}
\end{figure*}

Motivated by the above observations, we propose an \textbf{I}nstance-\textbf{A}ware \textbf{P}rompting \textbf{F}ramework (IAPF) tailored for training-free COS. As depicted in \Cref{fig: framework}, IAPF first uses an MLLM to generate image-specific sets of foreground and background category tags. Leveraging these text prompts, the proposed \textbf{Instance Mask Generator} instantiates a detector-agnostic enumerator to produce multiple instance-aware box prompts, addressing the core shortcomings of prior work. We further introduce a region-constrained point prompting strategy, Single-Foreground Multi-Background Prompting (SFMBP), which leverages multiple background tags to refine point prompts per instance. These carefully selected instance-level visual prompts (\ie, box-points pairs) enable SAM to produce precise, fine-grained candidate instance masks. Finally, we employ a Self-Consistency Instance Mask Voting mechanism over multiple candidates to yield robust and accurate COS results.

Our contributions are threefold:
\begin{enumerate}
\item We present IAPF that upgrades prompt granularity from semantic-level to instance-level for training-free COS, enabling accurate and complete segmentation under multi-instance camouflage.

\item Instance Mask Generator. We pair a detector-agnostic box enumerator (for proposing multiple instance-level box prompts) with SFMBP (for sampling instance-level point prompts), prompting SAM to accurately and completely produce instance masks in multi-instance camouflage.

\item Extensive experiments on three challenging COS benchmarks, two Camouflaged Instance Segmentation (CIS) benchmarks, and two downstream application datasets demonstrate that IAPF achieves SOTA performance among training-free methods.
\end{enumerate}

\section{Related Work}
\subsection{Camouflaged Object Segmentation}
Early fully supervised COS models~\cite{fan2020camouflaged,hu2023high} achieved strong accuracy with dense pixel masks, but the annotation cost is prohibitive, and every new domain typically requires re-annotation and retraining. Weakly supervised methods~\cite{chen2025just,he2023weakly,li2025weakly} reduce labeling by using sparse cues (points/scribbles/boxes), yet performance degrades as supervision becomes scarce and task-specific optimization remains necessary. Unsupervised pipelines~\cite{zhang2023unsupervised,shou2025sdalsnet,du2025shift} remove labels but exhibit a clear gap to fully supervised performance. Overall, these families are training-centric: their effectiveness scales with supervision and per-task fine-tuning. This motivates a training-free alternative that operates entirely with frozen models and zero task-specific labels, which be pursued by converting text cues into effective visual prompts at inference time.

\subsection{Segment Anything Model for COS}
SAM~\cite{kirillov2023segment} delivers accurate masks when supplied with explicit visual prompts (points/boxes). COS adaptations fall into two lines: (i) training-based variants that fine-tune SAM with dense~\cite{chen2023sam,gao2025cod} or sparse~\cite{he2023weaklysam,chen2024sam} supervision, and (ii) training-free pipelines that synthesize prompts at inference stage. Representative training-free approaches include: GenSAM~\cite{hu2024relax}, which uses an MLLM to produce text tags and derives visual prompts via CLIP heatmaps; ProMaC~\cite{hu2024leveraging}, which exploits hallucination priors to stabilize MLLM-generated tags; UpGen~\cite{du2025upgen}, which introduces generative priors to enhance training-free localization and prompt formation; INT~\cite{hu2025int}, which performs negative mining to improve the discriminativeness of task-generic promptable segmentation; and RDVP-MSD~\cite{yin2025stepwise}, which mitigates noisy heatmaps by sampling points within a single region-constrained box. Despite these advances, the existing training-free methods remain at the semantic-level (global or single-box prompts), lacking explicit instance enumeration—struggling when an image contains multiple camouflaged objects.

\section{Method}

\subsection{Framework Overview}
We propose the Instance-Aware Prompting Framework (IAPF), a training-free pipeline for COS that upgrades prompt granularity from semantic to instance-level while keeping all components frozen (no task-specific training). Given a test image $X$ and a task-generic prompt $p_i^{g}$ (\eg, ``\textit{camouflaged animal}''), IAPF repeats the pipeline $I$ times with synonymous prompts $p_i^g (i=1,...,I)$ and proceeds in three frozen stages.
(1) Text Prompt Generator. An MLLM answers a small set of text queries derived from $p_i^{g}$ (and its synonymous variants) and returns image-specific category tags: a set of foreground tags $tag^{fg}$ and a set of background tags $tag^{bg}$.
\textbf{(2) Instance Mask Generator.} A detector-agnostic box enumerator converts $tag^{fg}_i$ into $N$ instance-level boxes prompts $\mathcal{B}_i=\{\mathcal{B}_{i,n}\}_{n=1}^{N}$. Conditioned on $tag^{fg}_i$ and $set(tag^{bg})$, the Single-Foreground Multi-Background Prompting (SFMBP) strategy derives region-constrained foreground/background point prompts inside each box using CLIP-based heatmaps, yielding per-box points $\{p^{fg}_{i,n},p^{bg}_{i,n}\}$. All box-points pairs $\langle \mathcal{B}_{i,n},\,p^{fg}_{i,n},\,p^{bg}_{i,n}\rangle (n=1, 2, ..., N)$ form $N$ instance-level visual prompts for SAM, which produces a candidate instance mask for the image.
(3) Self-consistency Instance Mask Voting. Running the above with $I$ times, yields multiple candidate instance masks; the final COS result is chosen as the one closest to the set mean.

\subsection{Text Prompt Generator}
Text Prompt Generator employs a frozen Multimodal Large Language Model (MLLM)~\cite{liu2024improved} to convert a synonymous task-generic prompt $p_i^{g}$ (repetition index $i\in\{1, \dots, I\}$) into image-specific category tags tailored for each input image. Formally, given an image $X\in\mathbb{R}^{H\times W\times 3}$ and a textual query $Q$, the MLLM generates a response sequence $y$ in an auto-regressive manner:
\begin{align}
p(y \mid X, Q; \theta) &= \prod_{t=1}^{T} p(y_t \mid y_{<t}, X, Q; \theta),
\end{align}
where $y_t$ is the token at step $t$, $y_{<t}$ are previously generated tokens, $T$ is the response length, and $\theta$ denotes the (frozen) model parameters.

Following the GenSAM protocol~\cite{hu2024relax}, for each $p^g_i$ we first obtain an image caption $C_i$, and then sequentially query the frozen MLLM with two textual queries $Q_i^{f}$ and $Q_i^{b}$:
\begin{align}
    tag_i^{fg} =& \text{MLLM} (X, C_i, Q^f_i),\\
    tag_i^{bg} =& \text{MLLM} (X, C_i, Q^f_i, tag_i^{fg}, Q^b_i),
\end{align}
where $Q_i^{f}:$ ``Name of the $p^g_i$ in one word.'' and $Q_i^{b}:$ ``Name of the environment of the $p^g_i$ in one word.'' The MLLM's single-word answers are recorded as the foreground tag $tag_i^{fg}$ and background tag $tag_i^{bg}$, respectively. The pair $(tag_i^{fg},\,\mathrm{set}(tag^{bg}))$ constitutes the image-specific category tags consumed by the Instance Mask Generator at repetition $i$.

\subsection{Instance Mask Generator}
Existing training-free COS pipelines~\cite{yin2025stepwise,hu2024relax,hu2024leveraging,tang2024chain} typically operate with a single semantic-level prompt (often one box per image), which constrains SAM~\cite{kirillov2023segment} to coarse semantic masks and fails on multi-instance camouflage. \textbf{Instance Mask Generator (IMG)} bridges the image-specific category tags from the Text Prompt Generator to instance-level visual prompts under the fully frozen setting. For the $i$-th repetition with tags $(tag^{fg}_i,\,\mathrm{set}(tag^{bg}))$, IMG performs: (1) \emph{Box prompting}—a detector-agnostic enumerator converts $tag^{fg}_i$ into instance-level box prompts $\mathcal{B}_i=\{\mathcal{B}_{i,n}\}_{n=1}^{N}$; (2) \emph{Point prompting}—conditioned on $tag^{fg}_i$ and $\mathrm{set}(tag^{bg})$, SFMBP samples region-constrained foreground/background point prompts inside each $\mathcal{B}_{i,n}$ using CLIP-based heatmaps, yielding per-box point prompts $\{p^{fg}_{i,n},\,p^{bg}_{i,n}\}$; (3) \emph{Mask prediction}—each box-points pair $\langle B_{i,n},\,p^{fg}_{i,n},\,p^{bg}_{i,n}\rangle$ is fed to SAM to produce an instance mask $m_{i,n}$, and the candidate semantic mask for repetition $i$ is $mask_i=\bigvee_{n=1}^{N} m_{i,n}$ for subsequent Self-Consistency Instance Mask Voting. This explicit instance-level guidance—multiple boxes with per-box points—enables robust segmentation of multiple camouflaged objects while keeping all components frozen (no task-specific training).

\subsubsection{Box Prompting}
Prior training-free COS pipelines~\cite{hu2024relax,hu2024leveraging,hu2025int,yin2025stepwise} often collapse to a single semantic box per image, which prevents SAM~\cite{kirillov2023segment} from producing true instance masks. We cast box prompting as a detector-agnostic, text-conditioned instance enumeration problem under frozen weights: given the image $X$ and the foreground tag $tag^{fg}_i$ from the Text Prompt Generator, a box enumerator converts these into multiple instance-level box prompts:
\begin{align}
\hat{\mathcal{B}}_i \;=\; \mathcal{E}_{\text{Detector}}(X,\,tag^{fg}_i),
\end{align}
where $\mathcal{E}_{\text{Detector}}$ is a plug-and-play open-vocabulary detector (instantiated with Grounding DINO~\cite{liu2024grounding} in our implementation). To prevent adjacent camouflaged objects from being merged, we apply a non-maximum suppression (NMS) with a deliberately low IoU threshold,
\begin{align}
\mathcal{B}_i \;=\; \mathrm{NMS}_{\text{IoU}=0.2}\!\left(\hat{\mathcal{B}}_i\right) \;=\; \{\mathcal{B}_{i,n}\}_{n=1}^{N},
\end{align}
where suppresses redundant overlaps while preserving nearby instances. This step upgrades the prompt granularity from semantic to instance by enumerating per-image boxes in a training-free manner, replaces the “single-box” assumption with detector-driven coverage of all camouflaged instances, and yields a measurable intermediate (box-prompt accuracy) to assess prompt quality, as shown in \Cref{fig: motivationTable}.

\subsubsection{Point Prompting}
\textbf{Motivation.} Prior training-free COS methods either draw points from image-level heatmaps~\cite{hu2024relax,hu2024leveraging} or confine sampling to one region-constrained (RC) box per image~\cite{yin2025stepwise}. Both choices are brittle for multi-instance camouflage: (i) backgrounds are heterogeneous (foliage/sand/rock, etc.), so a single background tag provides weak negatives and frequent confusion; (ii) the single-box setting collapses instances and prevents per-instance score calibration, so global thresholds either miss small objects or overfire around large ones. \emph{Semantic vs. instance level:} earlier methods yield semantic-level point prompts (global or single-region), whereas we explicitly generate instance-level point prompts per enumerated box to align supervision with each camouflaged instance.

\textbf{SFMBP (ours).} We introduce Single-Foreground Multi-Background Prompting, an instance-aware, multi-background scheme operating per enumerated box $\mathcal{B}_{i,n}$ at repetition $i$, using the foreground tag $tag^{fg}_i$ and multiple background tags $\text{set}(tag^{bg})$. Specifically, Spatial CLIP~\cite{hu2024relax} is used to generate a foreground heatmap $H^{f_i}$ and multiple background heatmaps $H^b = \{H^{b_1}, H^{b_2}, \dots, H^{b_L}\}$, each $H^{b_l}$ corresponds to a specific background tag $tag^{b_l}$ from  $\text{set}(tag^{bg})$,where $\text{set}(\cdot)$ denotes deduplication over the background tag list before heatmap computation, and $L$ is the length of $\text{set}(tag^{bg})$:
\begin{align}
H^{f_i} &= \text{Spatial CLIP}(X, tag_i^{fg}),\\
H^{b_l} &= \text{Spatial CLIP}(X, tag^{b_l}), \quad l = 1, 2, \dots, L
\end{align}
and we perform box-wise normalization to calibrate scores locally within each instance. High-confidence, per-box point prompts are then obtained by thresholding (optionally inside a slightly eroded box to avoid boundary ambiguity):
\begin{align}
    p_{i,n}^{fg} &= \left\{(x, y) \mid \mathrm{Norm}(H^{f_i} \mid \mathcal{B}_{i,n})_{(x,y)} \geq 0.9\right\},\\
    p_{i,n}^{\mathrm{bg}} &= \bigcup_{l=1}^L \left\{(x, y) \mid \mathrm{Norm}(H^{b_l} \mid \mathcal{B}_{i,n})_{(x,y)} \geq 0.9\right\},
\end{align}
where $\mathrm{Norm}(H \mid \mathcal{B}_{i,n})$ denotes normalization of heatmap $H$ within box $\mathcal{B}_{i,n}$ to $[0,1]$ and the union $\bigcup_{l=1}^{L}$ aggregates per-tag background candidates into a single instance-level background point prompts. SFMBP thus upgrades point prompting along two grounded axes: \emph{instance locality} (box-wise normalization/selection) and \emph{multi-background contrast} (aggregating cues from multiple backgrounds), yielding compact yet discriminative instance-level point prompts for SAM.

\begin{table*}[!ht] % * 表示这个表格在双栏模板中使用单栏
    
  \centering
  % \tabcolsep=0.9\linewidth
  \setlength{\tabcolsep}{4.5pt}
    \small
    \begin{tabular*}{\linewidth}{r|c|c|cccc|cccc|cccc}
    \toprule
         \multirow{2}{*}{Methods} & 
         \multirow{2}{*}{Venue} & 
         \multirow{2}{*}{Setting} & 
         \multicolumn{4}{c|}{NC4K} &
         \multicolumn{4}{c|}{COD10K-TEST} & 
         \multicolumn{4}{c}{CAMO-TEST}
         \\ 
         \cline{4-15}
         &
         &
         & $S_\alpha$↑ & $F_\beta^\omega$↑ & $M$↓ & $E_{m}^\phi$↑
         & $S_\alpha$↑ & $F_\beta^\omega$↑ & $M$↓ & $E_{m}^\phi$↑
         & $S_\alpha$↑ & $F_\beta^\omega$↑ & $M$↓ & $E_{m}^\phi$↑
        \\

    \midrule
    % \multicolumn{13}{c}{\textbf{Un-supervision Setting}}\\
    % \midrule
UCOS-DA \tiny \cite{zhang2023unsupervised} & ICCV23           & $U$                                            & .755   & .656   & .085   & .819   & .689   & .513   & .086   & .739   & .700   & .605   & .127   & .784   \\
SdalsNet  \tiny \cite{shou2025sdalsnet} & AAAI25          & $U$                                            & .739   & .642   & .085   & .824   & .697   & .525   & .072   & .780   & .697   & .601   & .117   & .799   \\
EASE \tiny \cite{du2025shift} & CVPR25            & $U$                                            & \textbf{.800}   & \textbf{.735}   & \textbf{.056}   & \textbf{.884}   & \textbf{.773}   & \textbf{.656}   & \textbf{.040}   & \textbf{.866}   & \textbf{.749}   & \textbf{.684}   & \textbf{.098}   & \textbf{.831}   \\

\midrule
    % \multicolumn{13}{c}{\textbf{Weak-supervision Setting}}\\
    % \midrule

WS-SAM  \tiny \cite{he2023weaklysam} & NeurIPS23         & $P$                                            & .813   & .734   & .057   & .859   & .790   & .663   & .039   & .856   & .718   & .602   & .102   & .757   \\
P-COD  \tiny \cite{chen2025just} & ECCV24       & $P$                                            & .822   & .748   & .051   & .889   & .784   & .650   & .042   & .859   & .798   & .727   & .074   & .872   \\
SAM-COD  \tiny \cite{chen2024sam}  & ECCV24           & $P$                                            & .858   & .802   & .041   & .918   & .831   & .725   & .031   & .901   & .820   & .760   & .066   & .885   \\
$^\times$BoxSAM  \tiny \cite{li2025weakly} & IVC25       & $P$                                            & .810   & -      & .069   & .823   & .808   & -      & .042   & .851   & .745   & -      & .102   & .749   \\

SS  \tiny \cite{zhang2020weakly} & CVPR20                & $S$                                            & .723   & .584   & .090   & .803   & .684   & .466   & .068   & .774   & .684   & .538   & .119   & .760   \\
SCWS  \tiny \cite{yu2021structure} & AAAI21              & $S$                                            & .757   & .659   & .072   & .846   & .708   & .535   & .059   & .813   & .719   & .616   & .105   & .810   \\
CRNet  \tiny \cite{he2023weakly} & AAAI23             & $S$                                            & .775   & .688   & .063   & .855   & .733   & .576   & .049   & .832   & .735   & .641   & .092   & .815   \\
WS-SAM  \tiny \cite{he2023weaklysam} & NeurIPS23                          & $S$                                            & .829   & .757   & .052   & .886   & .803   & .680   & .038   & .877   & .759   & .667   & .092   & .818   \\
MINet  \tiny \cite{niu2024minet} & ACM MM24     & $S$                                            & .793   & .709   & .061   & .869   & .749   & .596   & .049   & .840   & .750   & .669   & .091   & .840   \\
WSMD  \tiny \cite{zha2024weakly} & AAAI24              & $S$                                            & .797   & .700   & .064   & .868   & .761   & .600   & .049   & .839   & .793   & .704   & .079   & .866   \\
SAM-COD  \tiny \cite{chen2024sam} & ECCV24                          & $S$                                            & .859   & .795   & .039   & .912   & .833   & .728   & .029   & .904   & .836   & .779   & .060   & .903   \\
$^\times$BoxSAM  \tiny \cite{li2025weakly} & IVC25                          & $S$                                            & .860   & -      & .052   & .894   & .836   & -      & .044   & .883   & .830   & -      & .084   & .860   \\

SAM-COD  \tiny \cite{chen2024sam} & ECCV24                          & $B$                                            & .867   & \textbf{.813}   & \textbf{.037}   & .923   & .842   & \textbf{.745}   & .028   & .914   & .837   & \textbf{.786}   & .062   & .901   \\
$^\times$BoxSAM  \tiny \cite{li2025weakly} & IVC25                    & $B$                                            & \textbf{.877}   & -      & \textbf{.037}   & \textbf{.925}   & \textbf{.857}   & -      & \textbf{.027}   & \textbf{.919}   & \textbf{.859}   & -      & \textbf{.057}   & \textbf{.908}   \\

\midrule
    % \multicolumn{13}{c}{\textbf{Training-free Setting}}\\
    % \midrule
LLaVA1.5+SAM \tiny \cite{liu2024improved} & CVPR23 & $ZS$                                           & .609   & .543   & .253   & .676   & .653   & .551   & .193   & .719   & .571   & .526   & .292   & .631   \\
CLIP\_Surgey+SAM  \tiny \cite{li2025closer} & PR25 & $ZS$                                           & .712   & .626   & .152   & .755   & .661   & .531   & .174   & .699   & .671   & .601   & .179   & .729   \\
Grounded SAM  \tiny \cite{ren2024grounded} & Arxiv24      & $ZS$                                           & .771   & .721   & .135   & .806   & .745   & .667   & .146   & .784   & .683   & .654   & .203   & .723   \\
$^\times$MMCPF  \tiny \cite{tang2024chain} & ACM MM24     & $ZS$                                           & .768   & .681   & .082   & -      & .733   & .592   & .065   & -      & .749   & .680   & .100   & -      \\
GenSAM  \tiny \cite{hu2024relax} & AAAI24            & $ZS$                                           & .805   & .745   & .066   & .863   & .783   & .682   & .055   & .845   & .727   & .648   & .105   & .799   \\

ProMaC  \tiny \cite{hu2024leveraging} & NeurIPS24         & $ZS$                                           & .814   & .762   & .056   & .884   & .803   & .710   & .042   & .875   & .745   & .685   & .100   & .830   \\
$^\times$UpGen \tiny \cite{du2025upgen}  & TIP25        & $ZS$                                           & .838      & .788      & .054      & .887      & .806   & .708   & .051   & .856   & .779   & .718   & .091   & .833   \\
$^\times$INT \tiny \cite{hu2025int}  & IJCAI25        & ZS                                           & -      & -      & -      & -      & .808   & .722   & .037   & .883   & .772   & .734   & .086   & .853   \\
RDVP-MSD  \tiny \cite{yin2025stepwise} & ACM MM25         & $ZS$                                           & .823   & .764   & .056   & .873   & .825   & .743   & .038   & .877   & .796   & .750   & \textbf{.081}   & .848   \\
% ArgusCogito \tiny \cite{tan2025arguscogito}  & Arxiv25        & $ZS$                                           & -      & -      & -      & -      & .843   & .824   & .026   & .928   & .800   & .774   & .079   & .866   \\
\rowcolor{gray!20}
ours               &         & $ZS$                                           & \textbf{.862}   & \textbf{.828}   & \textbf{.043}   & \textbf{.916}   & \textbf{.856}   & \textbf{.799}   & \textbf{.033}   & \textbf{.917}   & \textbf{.803}   & \textbf{.768}   & \textbf{.081}   & \textbf{.864}   \\

    \bottomrule

  \end{tabular*}
  \caption{Quantitative comparison of COS methods across three benchmark datasets. The table compares state-of-the-art methods based on unsupervised, weakly-supervised, and training-free setups. '$U$' indicates unsupervised settings, while the weakly-supervised settings include '$P$' (point), '$S$' (scribble), and '$B$' (box). '$ZS$' represents zero-shot training-free approaches applied to COS. '$^\times$' denotes the unavailable code in the corresponding paper. '↑' indicates higher is better, and '↓' indicates lower is better. The best results of different settings are highlighted in \textbf{bold}.}
  \label{tab:sota_COS}
\end{table*}

\subsubsection{Mask Prediction}
Given the $i$-th repetition’s foreground tag $tag^{fg}_i$ and its enumerated boxes, SFMBP yields an instance-level prompt triplet for each detected instance, $\langle \mathcal{B}_{i,n},\,p^{fg}_{i,n},\,p^{bg}_{i,n}\rangle_{n=1}^{N}$—one box with its foreground and multi-background point prompts. Unlike prior semantic-level prompting that feeds SAM a single global/region cue, these per-instance triplets preserve instance identity and prevent mask bleeding across neighboring camouflaged objects. Each triplet is forwarded to the frozen SAM to obtain a camouflaged instance:
\begin{align}
    m_{i,n}= \text{SAM}(X \mid \langle \mathcal{B}_{i,n},\,p^{fg}_{i,n},\,p^{bg}_{i,n}\rangle), \, n=1,\dots,N
\end{align}
where $N$ is the number of instance-level box prompts associated with $tag^{fg}_i$. The resulting set $\{m_{i,1},\ldots ,m_{i,N}\}\subset \{0, 1\}^{H \times W}$ delineates the $N$ camouflaged instances discovered at repetition $i$.

To retain instance granularity for downstream use (\eg, per-instance analysis and subsequent self-consistency voting), we stack the masks along a new dimension to form a candidate instance mask:
\begin{align}
    mask_i = [m_{i,1}; m_{i,2}; \ldots; m_{i,N}],
\end{align}
with $mask_i \in \{0, 1\}^{N \times H \times W}$.

\subsection{Self-consistency Instance Mask Voting}
In the fully frozen setting, using synonymous prompts $p_i^g$ ($i=1,\dots,I$) leads to MLLM-induced variability: the frozen MLLM may return slightly different captions/tags across synonym choices and decoding passes (prompt sensitivity and stochastic decoding), which in turn perturbs the detector’s box prompts and the SFMBP's point prompts. For fair comparability across repetitions, we first project each candidate instance mask $mask_i\in\{0,1\}^{N\times H\times W}$ to a semantic mask $M_i\in\{0,1\}^{H\times W}$ by performing a per-pixel logical OR $\bigvee$ over the instance dimension $N$:
\begin{align}
M_i(x,y) \;=\; \bigvee_{n=1}^{N} mask_i(n,x,y).
\end{align}

Running the \textbf{pipeline} $I$ times (default $3$) produces $\{M_1,\dots,M_I\}$. We then select the repetition that is most self-consistent with the cohort:
\begin{align}
\hat{i} \;=\; \arg\min_{i}\; \left\|\, M_i \;-\; \frac{1}{I}\sum_{j=1}^{I} M_j \,\right\|_2,
\end{align}
where $\|\cdot\|_2$ denotes the pixel-wise $\ell_2$ distance over binary masks. Finally, $M_{\hat{i}}$ and the corresponding $mask_{\hat{i}}$ are reported as the COS and CIS outputs, respectively—voting is conducted on the semantic projection for comparability, while the selected index $\hat{i}$ returns the full instance-aware solution.

\section{Experiments}
\subsection{Experiment Settings}

\subsubsection{Datasets and Evaluation Metrics}
We evaluate COS on NC4K~\cite{lv2021simultaneously} (4,121), COD10K-TEST~\cite{fan2020camouflaged} (2,026), and CAMO-TEST~\cite{le2019anabranch} (250) using Structure-measure $S_{\alpha}$~\cite{fan2017structure}, weighted F-measure $F^{\omega}_{\beta}$, mean E-measure $E^{\phi}_{m}$, and MAE $M$ (higher is better except $M$). For CIS, we follow COCO-style metrics and report $AP$, $AP_{50}$, and $AP_{75}$ on NC4K and COD10K-TEST. For downstream applications, we additionally evaluate—still in the training-free setting—on ACOD-12K-TEST (1,492; concealed crop detection; RGB-D, using RGB only)~\cite{wang2024depth} and PlantCamo (1,250; plant camouflage)~\cite{Yang2025}, evaluated with the same COS metrics as above.

\subsubsection{Implementation details}
\label{sec: Implementation_details}
In our experiments, we employ LLaVA-1.5-13B~\cite{liu2024improved} as the MLLM. For the detector-agnostic box enumerator, we use the Grounding DINO~\cite{liu2024grounding}. For the CLIP model, we use the CLIP~\cite{radford2021learning} from the CS-ViT-L/14@336px variant. For the SAM, we deploy the HQ-SAM~\cite{ke2023segment} built on the ViT-H version. Our method is entirely train-free, requiring no fine-tuning. All experiments are conducted using the PyTorch framework, and we utilize two NVIDIA GeForce RTX 3090 GPUs with 24 GB of memory each for evaluation.

\subsection{Comparison with State-of-the-Art Methods}
\subsubsection{Quantitative and Qualitative Comparison}
We compare the proposed IAPF with unsupervised, weakly-supervised, and training-free methods on the NC4K, COD10K-TEST, and CAMO-TEST datasets, as shown in \Cref{tab:sota_COS}. In the zero-shot training-free setting, our method achieves significant improvements over all compared methods, including higher $S_{\alpha}$, $F_{\beta}^{\omega}$, and $E_m^{\phi}$ values. IAPF exceeds unsupervised baselines and, on NC4K and COD10K, even surpasses weakly supervised methods, all without any supervision or finetuning.

\begin{figure*}[t]
\centering
\includegraphics[width=1.0\textwidth]{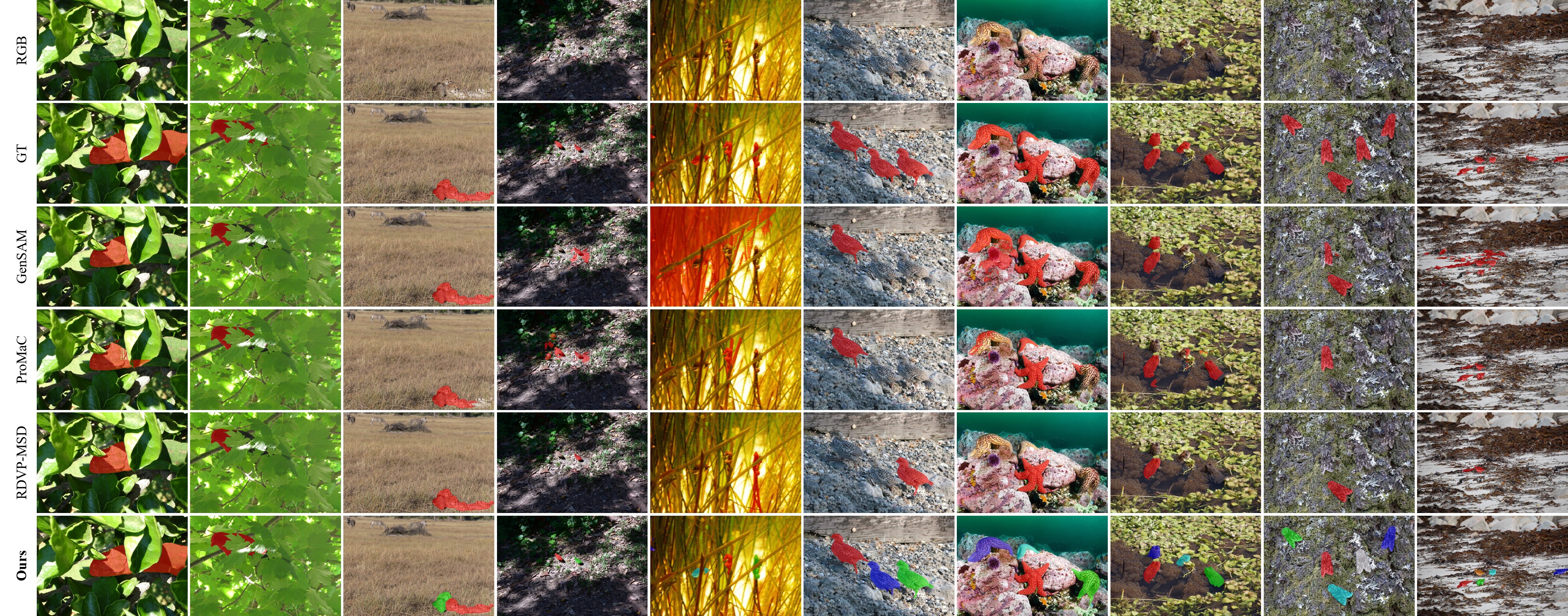} % Reduce the figure size so that it is slightly narrower than the column.
\caption{
Qualitative comparison of the proposed IAPF with three main training-free COS methods. From left to right, as the number of camouflaged instances in the scene increases, existing training-free methods fail to segment all objects. In contrast, the IAPF consistently produces high-quality instance masks, even when dealing with multiple camouflaged instances.
}
\label{fig: qualitativeComparison}
\end{figure*}

Qualitative results in \Cref{fig: qualitativeComparison} show that, as the number of camouflaged instances increases, previous training-free COS methods~\cite{hu2024relax,hu2024leveraging,yin2025stepwise} struggle to segment the objects accurately, while our method maintains strong performance across all scenarios.

\subsubsection{Evaluation on CIS datasets}
We further evaluate our method on the Camouflaged Instance Segmentation (CIS) task, reporting quantitative results on the NC4K and COD10K-TEST datasets in \Cref{tab:sota_CIS}. Compared to recent fully-supervised ('$F$') and image-level supervised ('$I$') CIS approaches, our method achieves competitive or even superior performance, notably attaining the highest $AP_{50}$ scores on both NC4K and COD10K-TEST datasets. Remarkably, our results are obtained entirely in a training-free manner using only a single task-generic prompt, underscoring the effectiveness of the instance-aware prompting strategy for accurate camouflaged instance segmentation without reliance on costly annotations.

\subsection{Ablation Study}
\subsubsection{Effectiveness of the Modules}
We perform ablation studies to examine the effectiveness of each key component in the IAPF framework: Text Prompt Generator (TPG), Grounding DINO (G), Single-Foreground Multi-Background Prompting (SFMBP), and Self-consistency Instance Mask Voting (SIMV). As shown in \Cref{tab:ablationForModules}, removing TPG (setting $1$), G and SFMBP (settings $2.1$-$2.3$), or SIMV (setting $3$) clearly degrades performance, verifying the necessity of each module. Moreover, comparisons between settings ($2.1$ \textit{vs.} $2.3$) and ($2.2$ \textit{vs.} $4$) further demonstrate that introducing Grounding DINO not only grants the IAPF the critical instance-awareness capability but also significantly improves overall segmentation accuracy by allowing the model to localize more camouflaged instances. The complete method (setting $4$) consistently outperforms all variants on both NC4K and COD10K-TEST datasets, confirming that each proposed component contributes distinctly and effectively to the final COS performance.

\subsubsection{Effectiveness of Repetition Number}
We evaluate the impact of the repetition number $I$ on the stability and performance of our method. As shown in \Cref{tab: repetitions}, increasing the number of repetitions improves the segmentation results. The best performance is achieved when $I = 3$, with higher repetitions providing diminishing returns.

\begin{table} % * 表示这个表格在双栏模板中使用单栏
    
  \centering
  %\tabcolsep=0.35cm    
  % \tabcolsep=0.005\linewidth
  \setlength{\tabcolsep}{1pt}
  \small
    \begin{tabular}{r|c|ccc|ccc}
    \toprule
         \multirow{2}{*}{Methods} &
         \multirow{2}{*}{Setting} & 
         \multicolumn{3}{c|}{NC4K} &
         \multicolumn{3}{c}{COD10K-TEST}
         \\ 
         \cline{3-8}
         &
         & $AP$ & $AP_{50}$ & $AP_{75}$
         & $AP$ & $AP_{50}$ & $AP_{75}$
        \\ 
\toprule
OSFormer \tiny \cite{pei2022osformer}        & $F$    & .444 & .737 & .451 & .420   & .713 & .428 \\
CLPNet \tiny \cite{cui2023contour}        & $F$    & .451 & .748 & .455 & .433   & .719 & .432 \\
QFormer \tiny \cite{dong2023unified}       & $F$    & .501 & .768 & .528 & .454   & .718 & .479 \\
DCNet \tiny \cite{luo2023camouflaged}             & $F$    & .528 & .771 & .565 & .453   & .707 & .475 \\
AQSFormer \tiny \cite{dong2024adaptive}      & $F$    & .505 & .768 & .535 & .465   & .738 & .485 \\
Mask2Camouflage \tiny \cite{phung2024revealing}  & $F$    & \textbf{.538} & .776 & \textbf{.583} & \textbf{.468}   & .725 & \textbf{.490} \\
TPNet \tiny \cite{he2024text}          & $I$ & .214 & .483 & .166 & .183   & .418 & .143 \\
\rowcolor{gray!20}
Ours                                           &$ZS$      & .511 & \textbf{.807} & .555 & .448   & \textbf{.748} & .475\\
    \bottomrule
  \end{tabular}
  \caption{Quantitative evaluation on CIS across NC4K and COD10K-TEST datasets. 'F' and 'I' denote fully-supervised and image-level supervised methods, respectively. ‘ZS’ denotes zero-shot training-free setting.
}
  \label{tab:sota_CIS}
\end{table}

\begin{table} % * 表示这个表格在双栏模板中使用单栏
    
  \centering
  %\tabcolsep=0.35cm
  % \tabcolsep=0.005\linewidth
  \setlength{\tabcolsep}{1pt}
  \small
    \begin{tabular}{l|cccc|cccc}
    \toprule
         \multirow{2}{*}{Method's Variants} &
         \multicolumn{4}{c|}{NC4K} &
         \multicolumn{4}{c}{COD10K-TEST}
         \\ 
         \cline{2-9}
         & $S_\alpha$↑ & $F_\beta^\omega$↑ & $M$↓ & $E_{m}^\phi$↑
         & $S_\alpha$↑ & $F_\beta^\omega$↑ & $M$↓ & $E_{m}^\phi$↑
        \\ 
\toprule
(1) \textit{wo} TPG& .855   & .816   & .046   & .909&.840    &.768    &.036    &.902       \\
(2.1) \textit{wo} G\&\textit{wo} SFMBP& .788   & .707   & .074   & .840& .760   & .626   & .064   & .815      \\
(2.2) \textit{wo} G\&\textit{w} SFMBP&.797    &.714    &.062    & .850&.766    &.627    &.053    &.822       \\
(2.3) \textit{w} G\&\textit{wo} SFMBP&.847    &.802    &.051    &.900&.835    &.760    &.041    &.894        \\
(3) \textit{wo} SIMV&.850          & .812          & \textbf{.042}          & .902          & .844          & .780          & \textbf{.031}          & .904          \\
\rowcolor{gray!20}
(4) Ours& \textbf{.862}   & \textbf{.828}   & .043   & \textbf{.916}&\textbf{.856}   & \textbf{.799}   & .033   & \textbf{.917}      \\
    \bottomrule
  \end{tabular}
  % \caption{Ablation study on the effectiveness of IAPF components.}
  \caption{Ablation study on the effectiveness of the IAPF.}
  \label{tab:ablationForModules}
\end{table}

\begin{table} % * 表示这个表格在双栏模板中使用单栏
    
  \centering
  \tabcolsep=0.018\linewidth
  \small
    \begin{tabular}{c|cccc|cccc}
    \toprule
         \multirow{2}{*}{Repeat} &
         \multicolumn{4}{c|}{NC4K} &
         \multicolumn{4}{c}{COD10K-TEST}
         \\ 
         \cline{2-9}
         & $S_\alpha$↑ & $F_\beta^\omega$↑ & $M$↓ & $E_{m}^\phi$↑
         & $S_\alpha$↑ & $F_\beta^\omega$↑ & $M$↓ & $E_{m}^\phi$↑
        \\ 
\toprule
1      & .850          & .812          & .042          & .902          & .844          & .780          & .031          & .904          \\
2      & .857          & .820          & .045          & .912          & .852          & .789          & .034          & .913          \\
\rowcolor{gray!20}
3      & \textbf{.862} & \textbf{.828} & \textbf{.043} & \textbf{.916} & \textbf{.856} & \textbf{.799} & \textbf{.033} & \textbf{.917} \\
4      & .861          & .827          & .044          & .914          & .855          & .798          & \textbf{.033} & .915          \\
5      & .859          & .824          & .044          & .912          & .855          & .796          & .034          & .914          \\
    \bottomrule
  \end{tabular}
  \caption{Ablation study on the effectiveness of repetition number $I$.}
  \label{tab: repetitions}
\end{table}

\subsubsection{Performance Comparison of Different Camouflaged Instance Counts}
\Cref{tab: instanceDifferent} reports COS performance on COD10K-TEST after stratifying images by the number of camouflaged instances: (1) single, (2) double, and (3) three-or-more objects. For each metric, we also list the relative change $(\%)$ \textit{w.r.t.} the single-instance baseline (row $1$ of every method).

The proposed IAPF maintains consistently high accuracy as instance density grows. The larger percentage swing in $M$ for all methods is expected because the absolute MAE values are very small; nonetheless, the absolute MAE of the IAPF remains the lowest. These results confirm that the instance-aware prompting pipeline––particularly the detector-agnostic box enumerator–based box prompts and SFMBP point sampling––provides strong resilience to cluttered scenes with many camouflaged objects.

\subsubsection{Comparison With Other SOTAs on a Fair Basis}
To ensure a fair comparison, we evaluate all methods using the same model configuration, including LLaVA-1.5-13B as the MLLM, CS-ViT-L/14@336px for CLIP, and HQ-SAM built on ViT-H, as mentioned in \Cref{sec: Implementation_details}. The results in \Cref{tab:fairComparison} demonstrate that our method outperforms GenSAM~\cite{hu2024relax} and ProMaC~\cite{hu2024leveraging} by a significant margin. RDVP-MSD~\cite{yin2025stepwise}, which maintains the same configuration as our approach, shows no performance improvement, confirming the stability of our setup. Even with identical model configurations, our method consistently leads in all metrics, clearly showcasing its superior ability to handle multi-instance camouflage and produce accurate, training-free segmentation.

\subsection{Downstream Applications}
\begin{figure}[t]
  \centering
   \includegraphics[width=1.0\linewidth]{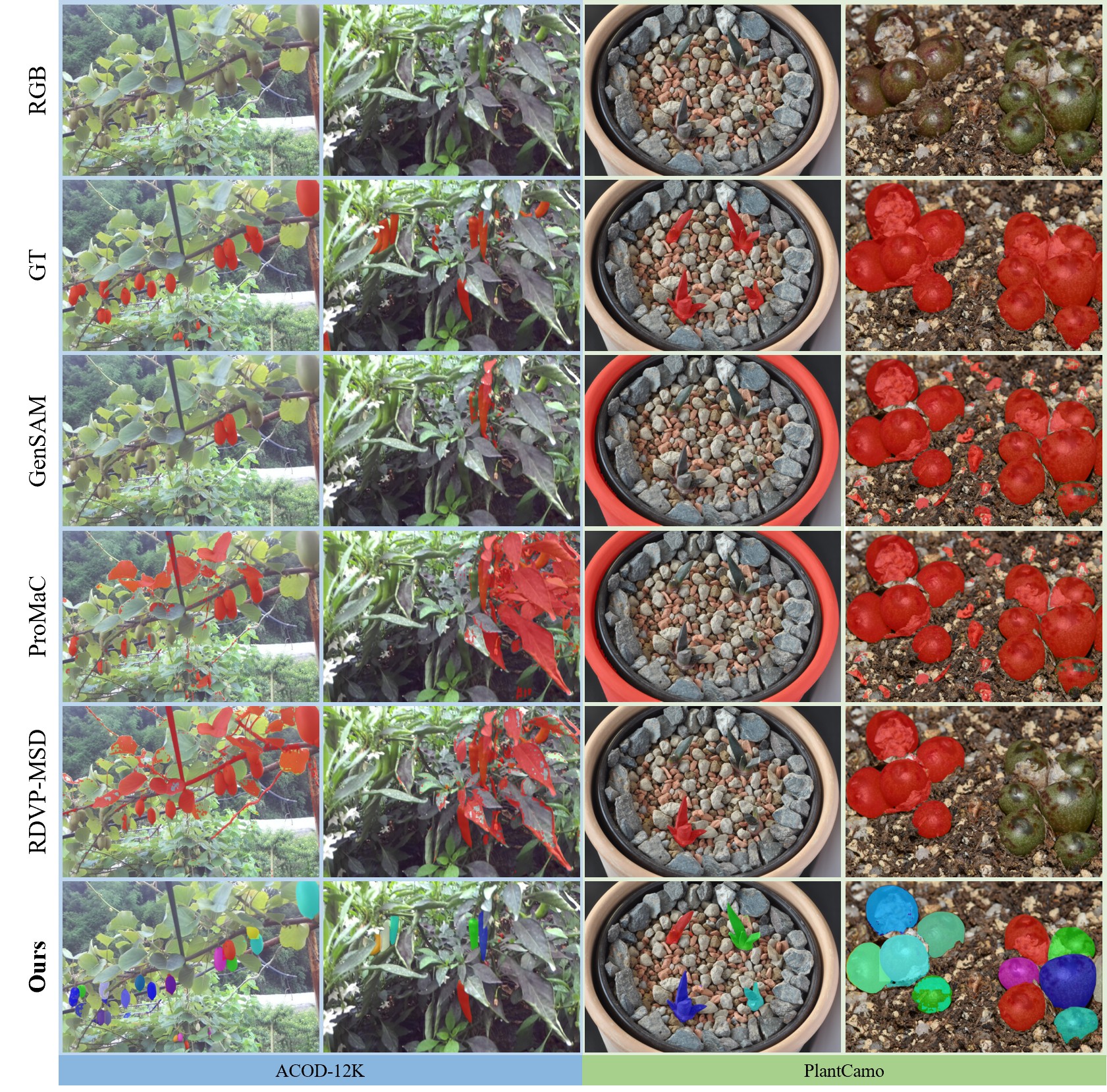}
\caption{Downstream qualitative results on ACOD-12K and PlantCamo.}
\label{fig: qualitativeDownstream}
\end{figure}

We evaluate on two real-world, multi-instance datasets—ACOD-12K (dense agricultural crop, RGB-only) and PlantCamo (plant camouflage). In these scenes, targets are small and dispersed; other training-free baselines that feed SAM semantic-level prompts collapse multiple objects or leak to textured backgrounds, while IAPF preserves instance separation (\Cref{fig: qualitativeDownstream}). Quantitatively, IAPF achieves the best results across all metrics on both datasets, with clear margins (see~\Cref{tab:application}), demonstrating that upgrading prompts from semantic-level to instance-level is crucial for robust generalization to downstream tasks with multiple scattered objects.

\begin{table} % * 表示这个表格在双栏模板中使用单栏
    
  \centering
  \tabcolsep=0.010\linewidth
  \small
    \begin{tabular}{c|c|llll}
    \toprule
         \multirow{2}{*}{Methods} &
         \multirow{2}{*}{Ins} &
         \multicolumn{4}{c}{COD10K-TEST}
         \\ 
         \cline{3-6}
         & & $S_\alpha$↑ & $F_\beta^\omega$↑ & $M$↓ & $E_{m}^\phi$↑
        \\ 
\toprule
   & (1) & .789 & .690 & .055 & .852 \\
GenSAM                          & (2) & .730 \tiny -7.48\% & .622 \tiny -9.86\% & .064 \tiny -16.36\% & .820 \tiny -3.76\% \\
                          & (3) & .632 \tiny -19.90\% & .443 \tiny -35.80\% & .077 \tiny -40.00\% & .700 \tiny -17.84\% \\
\midrule
   & (1) & .814 & .725 & .040 & .881 \\
ProMaC                          & (2) & .760 \tiny -6.63\% & .667 \tiny -8.00\% & .051 \tiny -27.50\% & .855 \tiny -2.95\% \\
                          & (3) & .634 \tiny -22.11\% & .457 \tiny -36.97\% & .073 \tiny -82.50\% & .750 \tiny -14.87\% \\
\midrule
 & (1) & .833 & .756 & .038 & .886 \\
RDVP-MSD                          & (2) & .774 \tiny -7.08\% & .674 \tiny -10.85\% & .042 \tiny -10.53\% & .834 \tiny -5.87\% \\
                          & (3) & .648 \tiny -22.21\% & .455 \tiny -39.81\% & .079 \tiny -107.89\% & .672 \tiny -24.15\% \\
\midrule
\rowcolor{gray!20} 
     & (1) & \textbf{.860} & \textbf{.804} & \textbf{.033} & \textbf{.918} \\ 
\rowcolor{gray!20} 
Ours                          & (2) & \textbf{.840} \tiny -2.33\% & \textbf{.788} \tiny -1.99\% & \textbf{.032} \tiny +3.03\% & \textbf{.921} \tiny +0.33\% \\
\rowcolor{gray!20} 
                          & (3) & \textbf{.777} \tiny -9.65\% & \textbf{.697} \tiny -13.31\% & \textbf{.050} \tiny -51.52\% & \textbf{.846} \tiny -7.84\% \\

    \bottomrule
  \end{tabular}
  \caption{Performance comparison of different camouflaged instance counts on COD10K-TEST.}
  \label{tab: instanceDifferent}
\end{table}

\begin{table} % * 表示这个表格在双栏模板中使用单栏
    
  \centering
  %\tabcolsep=0.35cm
  % \tabcolsep=0.005\linewidth
  \setlength{\tabcolsep}{3pt}
  \small
    \begin{tabular}{c|cccc|cccc}
    \toprule
         \multirow{2}{*}{Methods} &
         \multicolumn{4}{c|}{NC4K} &
         \multicolumn{4}{c}{COD10K-TEST}
         \\ 
         \cline{2-9}
         & $S_\alpha$↑ & $F_\beta^\omega$↑ & $M$↓ & $E_{m}^\phi$↑
         & $S_\alpha$↑ & $F_\beta^\omega$↑ & $M$↓ & $E_{m}^\phi$↑
        \\ 
\toprule
GenSAM& .827   & .774   & .054   & .880&.811    &.721    &.041    &.867       \\
ProMaC& .826   & .779   & .053   & .890&.818    &.738    &.038    &.887       \\
RDVP-MSD& .823   & .764   & .056   & .873&.825    &.743    &.038    &.877       \\
\rowcolor{gray!20}
Ours& \textbf{.862}   & \textbf{.828}   & \textbf{.043}   & \textbf{.916}&\textbf{.856}   & \textbf{.799}   & \textbf{.033}   & \textbf{.917}      \\
    \bottomrule
  \end{tabular}
  \caption{The fair comparison between the proposed method and other SOTA methods.}
  \label{tab:fairComparison}
\end{table}

\begin{table} % * 表示这个表格在双栏模板中使用单栏
    
  \centering
  %\tabcolsep=0.35cm
  % \tabcolsep=0.005\linewidth
  \setlength{\tabcolsep}{3pt}
  \small
    \begin{tabular}{c|cccc|cccc}
    \toprule
         \multirow{2}{*}{Methods} & 
         \multicolumn{4}{c|}{ACOD-12K} &
         \multicolumn{4}{c}{PlantCamo}
         \\ 
         \cline{2-9}
         & $S_\alpha$↑ & $F_\beta^\omega$↑ & $M$↓ & $E_{m}^\phi$↑
         & $S_\alpha$↑ & $F_\beta^\omega$↑ & $M$↓ & $E_{m}^\phi$↑
        \\

    \midrule
    % \multicolumn{13}{c}{\textbf{Un-supervision Setting}}\\
    % \midrule

GenSAM & .478   & .202   & .193   & .517   & .670   & .540   & .114   & .743      \\
ProMaC & .457   & .175   & .225   & .499   & .651   & .521   & .124   & .757      \\
RDVP-MSD & .453   & .147   & .202   & .505   & .704   & .582   & .100   & .761      \\

\rowcolor{gray!20}
ours               & \textbf{.603}   & \textbf{.406}   & \textbf{.103}   & \textbf{.660}   & \textbf{.758}   & \textbf{.664}   & \textbf{.086}   & \textbf{.829}      \\

    \bottomrule

  \end{tabular}
  \caption{Downstream quantitative results on ACOD-12K and PlantCamo.}
  \label{tab:application}
\end{table}

\section{Conclusion}
We presented IAPF, the first training-free COS framework that explicitly converts a single task-generic prompt into instance-level visual prompts—multiple boxes plus per-box foreground/multi-background points—for SAM. A frozen MLLM yields image-specific category tags; a detector-agnostic enumerator produces instance boxes; SFMBP derives contrastive per-box points; and SIMV selects the most self-consistent instance solution. These upgrades prompt granularity from semantic to instance-level and enable fine-grained, multi-instance COS without any task-specific training or annotations. Extensive results on three COS benchmarks, two CIS benchmarks, and two downstream datasets demonstrate state-of-the-art performance among training-free methods and strong cross-domain generalization, establishing instance-aware prompting as a simple yet powerful paradigm for promptable segmentation.
{
    \small
    \bibliographystyle{ieeenat_fullname}
    \bibliography{main}
}


\end{document}